\documentclass[10pt,twocolumn,letterpaper]{article}

\usepackage{cvpr}
\usepackage{times}
\usepackage{epsfig}
\usepackage{graphicx}
\usepackage{amsmath}
\usepackage{amssymb}
\usepackage{multirow}

\usepackage[pagebackref=true,breaklinks=true,letterpaper=true,colorlinks,bookmarks=false]{hyperref}

\cvprfinalcopy

\def\cvprPaperID{*} 

\ifcvprfinal\fi
\begin{document}

\title{T-VSE: Transformer-Based Visual Semantic Embedding}

\author{Muhammet Bastan\\
    Amazon\\
    Palo Alto, CA, USA\\
    {\tt\small mbastan@amazon.com}
\and
    Arnau Ramisa\\
    Amazon\\
	Palo Alto, CA, USA\\
	{\tt\small aramisay@amazon.com}
\and
    Mehmet Tek\\
    Google\\
    Mountain View, CA, USA\\
    {\tt\small mtek@google.com}
}

\maketitle

\begin{abstract} 
	Transformer models have recently achieved impressive performance on NLP tasks, owing to new algorithms for self-supervised pre-training on very large text corpora.  In contrast, recent literature suggests that simple average word models outperform more complicated language models, e.g., RNNs and Transformers, on cross-modal image/text search tasks on standard benchmarks, like MS COCO.
	In this paper, we show that dataset scale and training strategy are critical and demonstrate that transformer-based cross-modal embeddings outperform word average and RNN-based embeddings by a large margin, when trained on a large dataset of e-commerce product image-title pairs. 	
\end{abstract}

\section{Introduction}
\begin{figure}
	\centering	
	\includegraphics[width=0.35\textwidth]{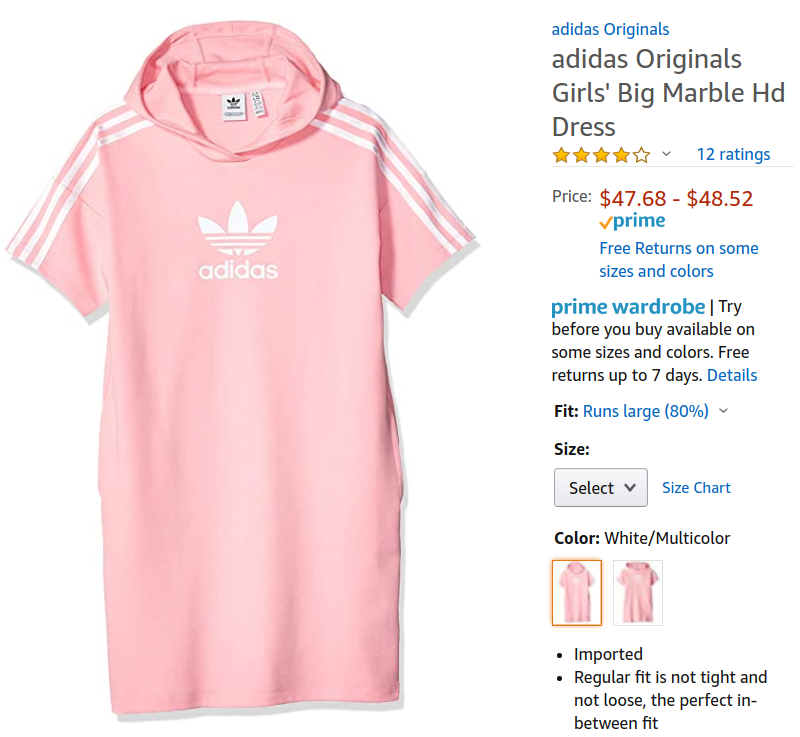}
	\caption{E-commerce sites have huge amounts of image-text pairs which can be used to learn cross-modal representations. The image is from amazon.com.}
	\label{fig:asin}
\end{figure}

Cross-modal representation learning can leverage the huge amounts of multi-modal image-text data (Fig.~\ref{fig:asin}) that is readily available on e-commerce sites. Each product has an image, a title as a brief description of the product and, often, additional complementary text metadata. Cross-modal learning can utilize this multi-modal data to learn a good representation of the product, which can be used for cross-modal (text-to-image, image-to-text) product search, clustering, de-duplication, recommendation, etc.

Visual semantic embedding (VSE)~\cite{vse-bmvc18} uses (image, text) pairs to learn a low-dimensional common embedding space (Fig.~\ref{fig:vse}). VSE models typically consist of two-stream neural networks (NN), one CNN branch to encode images and one NN branch to encode the text (Fig.~\ref{fig:vse}). The text consists of a sequence of tokens and requires sequence modeling. Several different models have been employed for text encoding in the literature: RNNs (LSTM/GRU), variants of word2vec, simple word averaging, and more recently \textbf{transformers}.

Transformer models~\cite{transfomer-nips17,bert-naacl19,xlnet-nips19} have recently achieved state-of-the-art performance on NLP tasks\footnote{https://gluebenchmark.com} and replaced the previously popular RNN models (LSTM/GRU). This success is due mostly to the self-supervised pre-training of the transformer models on very large text datasets and then fine-tuning on target tasks.

In contrast to their success in text encoding for NLP tasks, transformers have not been shown to work well for cross-modal vision-language tasks, such as VSE. In fact, a recent work~\cite{lang-iccv19} reported results claiming ``an average embedding language model outperforms an LSTM on retrieval-style tasks; state-of-the-art representations such as BERT perform relatively poorly on vision-language tasks.'' 

In this paper, contrary to~\cite{lang-iccv19}, we show that transformer-based VSE (T-VSE) actually works much better than word average and RNN-based VSE models. The key to the success of T-VSE is properly training it on a \textbf{large} dataset, whereas the standard VSE datasets are relatively small, e.g., MS COCO has 128K (image, caption) pairs. To this end, we constructed a large dataset of 12.1M (image, title) pairs from fashion items listed on amazon.com.

\section{Visual Semantic Embedding (VSE)}
\label{sec:vse}
\begin{figure}
	\centering	
	\includegraphics[width=0.40\textwidth]{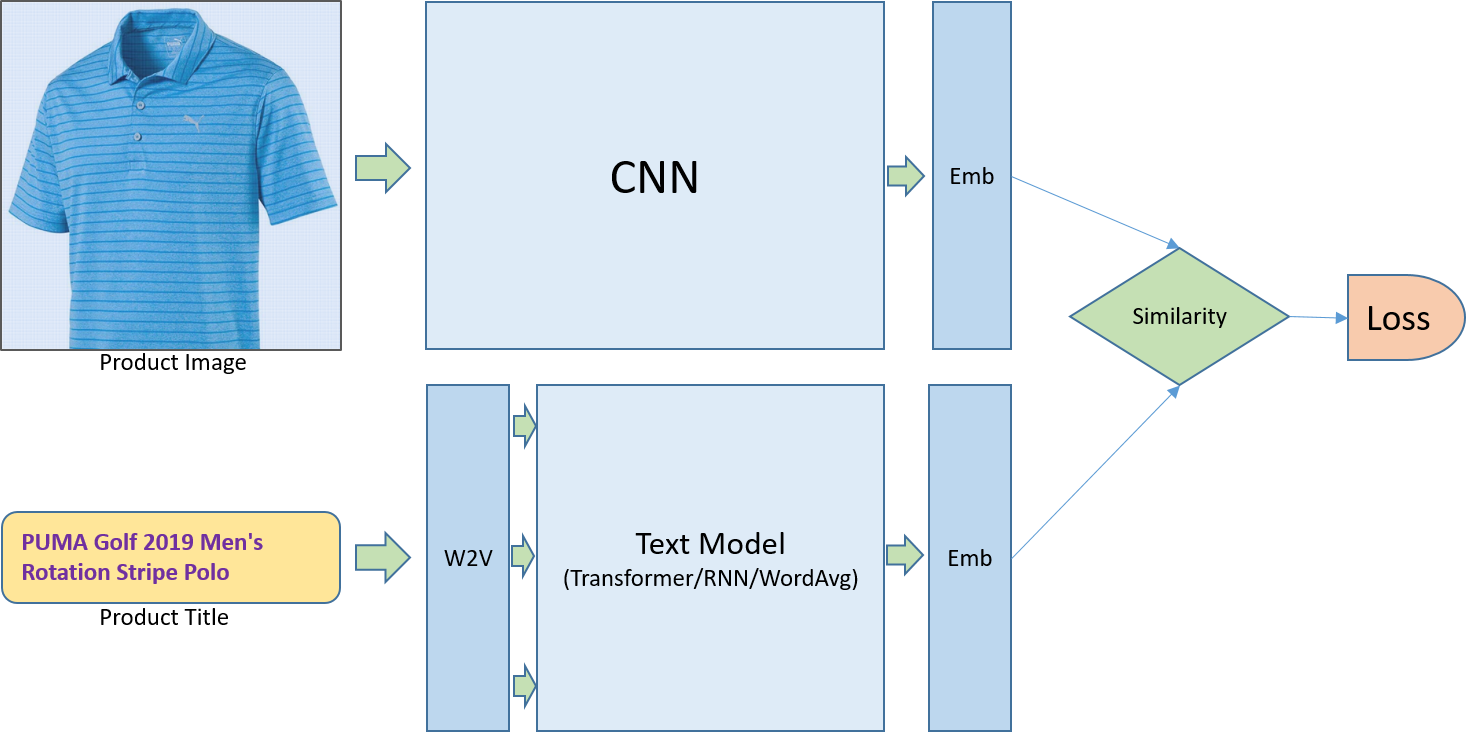}
	\caption{Visual semantic embedding (VSE) learning framework used in this paper.}
	\label{fig:vse}
\end{figure}

We use the classical VSE framework, shown in Fig.~\ref{fig:vse} for cross-modal retrieval (text-to-image and image-to-text).
Given pairs of images and their text descriptions $(i_k, t_k)$, VSE learns embedding functions, or encoders, $f(i_k)$ and $g(t_k)$, by maximizing the similarity $s(i_k, t_k)$ between the positive (image, text) pairs, while minimizing the similarity between negative pairs $(i_k, t_j)$, $k \ne j$. The encoders $f$ and $g$ are typically neural networks (CNNs for images, and RNNs, MLPs or transformers for text), and the similarity function $s$ is the cosine similarity. The VSE model can be trained by optimizing a suitable metric learning loss function, such as the contrastive or the triplet loss.

\subsection{Loss Function}
\label{sec:loss}
In~\cite{vse-bmvc18}, the authors proposed the \textit{Max of Hinges (MH)} loss function, which has proved to be very effective in training VSE models. MH loss is basically a symmetric triplet loss with hard negative mining. Given a batch of $N$ (image, text) pairs $(i_k, t_k)$, image and text encoders $f$ and $g$, and a similarity function $s$ (inner product), the loss function is defined as two symmetric terms, one for image-to-text and another for text-to-image:
\begin{equation}
\begin{split} 
L &= \sum_{k \ne j}^{N}  [\max_j s( f(i_k), g(t_j) ) - s( f(i_k), g(t_k) ) + m]_+\\
  & + \sum_{k \ne j}^{N} [\max_j s(  f(i_j), g(t_k) ) - s( f(i_k), g(t_k) ) + m]_+
\end{split}
\end{equation}
where $m$ is a margin value, and $[x]_+ \equiv max(x, 0)$. Hard negative mining is used in both terms, which can be problematic with the presence of duplicates or near-duplicates, as they will be treated as hard negatives. However, the probability of duplicates falling in the same batch decreases as the dataset size increases, and it is also possible to mitigate the problem during sampling and by re-weighting the loss.

\subsection{Image and Text Encoders}

As image encoder $f$, we used DenseNet 169, replacing the final classification layer with a linear embedding layer of dimensionality $D=256$. Higher $D$ gives slightly better accuracy, at the expense of higher computational cost.

Our main focus in this paper is transformer-based VSE models, i.e., the text encoder $g$ uses a transformer model. We also experimented with two other widely used text models for comparison, as in~\cite{lang-iccv19}: Word Average and RNN. All three models use the same word embedding layer (W2V in Fig.~\ref{fig:vse}), that projects one-hot encoded text token vectors to word embeddings of low-dimensionality, which in turn are fed to the text model. Finally, a linear embedding layer of size $D=256$ computes the embedding for the whole text sequence (Fig.~\ref{fig:vse}).

\noindent
\textbf{Word Average Model (AVG-VSE).} This is the simplest model, taking the word embeddings as input and computing their average as the output. Hence, this model discards the positional information of the input tokens. We inserted an additional fully connected layer of output size $512$ before the final embedding layer, as in~\cite{lang-iccv19}.

\noindent
\textbf{RNN Model (RNN-VSE).} RNN models have been mainstream in VSE~\cite{vse-bmvc18}. LSTM and GRU are widely used to capture the sequential nature of text.  We used a two-layer, unidirectional GRU, and fed the output of the last hidden state to the final embedding layer. We also experimented with a bidirectional GRU, and with taking the mean of the hidden states, but did not observe any significant difference in performance. A major drawback of RNN models is that they process the input sequentially and are, therefore, not parallelizable.

\noindent
\textbf{Transformer Model (T-VSE).} Although they were first proposed in~\cite{transfomer-nips17}, transformer models gained popularity with the BERT model~\cite{bert-naacl19} which, with the help of pre-training on large unlabeled text data and fine-tuning on target tasks, achieved state-of-the-art results on NLP benchmarks. Many variants of BERT have since been proposed, with slight modifications in the architecture and/or the pre-training algorithm. XLNet~\cite{xlnet-nips19}, RoBERTa~\cite{roberta-2019}, ALBERT~\cite{albert-2019} are only few of them.

Transformers are large --but parallelizable-- feed-forward networks, that include self-attention (inner products and softmax), linear, and normalization layers. They can learn context very well due to the self-attention mechanism, and can include a positional encoding layer, whose output is added to the word embeddings and fed to the transformer layers to take into account the ordering of the words. Because of their size, self-supervised pre-training on large unlabeled datasets is key to the success of transformers.

In this paper, we used the DistilBERT model~\cite{distilbert-2019}, a lightweight BERT model with $6$ transformer layers, which is 40\% smaller and 60\% faster than the original BERT-base model. 
The DistilBERT model in \cite{distilbert-2019} was trained by knowledge distillation with BERT-base as the teacher network, achieving 97\% of its performance. We only use the network architecture and train it from scratch as described below, without knowledge distillation. We also experimented with a 12-layer DistilBERT model, which has almost the same size as the original BERT base model.

Transformer models typically use a maximum sequence length of $512$ for NLP tasks, which is too long for our task of product image-title embeddings. Since the average title length is $17\pm5$, we follow the \textit{three-sigma} rule and set a maximum sequence length of $32$, which leads to considerable memory and computation savings.

\subsection{Dataset}
We constructed a new large scale dataset, \textbf{Amazon Fashion 12M} (AF12M)\footnote{We are planning to release the AF12M dataset.}, consisting of about 12.1M (product image, product title) pairs from amazon.com in US (Fig.~\ref{fig:asin}). The (image, title) pairs are readily available and no annotation is required.

The dataset was split as follows: 11M (image, title) pairs for training, 100K for validation, and 1M for testing.

\subsection{Text Preprocessing and Tokenization}
To prepare the product titles for the language model, we first applied Unicode NFKC normalization, followed by ASCII encoding. Next, we removed special characters, corrected common typos, and tokenized the titles according to a vocabulary built from the training set.

The tokenization algorithm and vocabulary size have a significant effect on the network size. For example, \cite{semsearch-kdd19} used a very large vocabulary (500K), consisting of word unigrams, bigrams, character trigrams and out-of-vocabulary bins, which required large computational resources to train even simple text models.  On the other hand, recent state-of-the-art NLP models employ either word-piece or byte-pair encoding~\cite{unigram-acl18, bpe-acl15} tokenization algorithms, resulting in much smaller vocabularies (30K, 50K) \cite{roberta-2019, distilbert-2019} and, in turn, smaller and more efficient networks. These tokenization algorithms consider text as a sequence of bytes and are language agnostic. They keep the frequent words as they are while representing rare words with sub-word units, which solves the out-of-vocabulary problem. Based on these insights, we decided to use word-piece tokenization.

The pre-trained transformer models\footnote{https://github.com/huggingface/transformers} come with vocabularies trained on generic text, but because of the highly specialized nature of the text in our dataset, we found that using vocabularies learned from our training set leads to better results. We used the SentencePiece\footnote{\label{fn-sp}https://github.com/google/sentencepiece} tokenizer's~\cite{sp-emnlp18} `unigram' model~\cite{unigram-acl18}, which is actually a sub-word tokenizer. We also tried the BPE~\cite{bpe-acl15} tokenizer, but it generated less meaningful and more sparse vocabularies, which negatively affects model performance.

We trained vocabularies of size 10K, 20K, 30K and 40K. Larger vocabulary sizes translate to slightly higher accuracy, in exchange for an increased memory and computational cost. A vocabulary of 30K is a good trade-off.

\subsection{Training}
\label{sec:train}
We trained all three VSE models (AVG-VSE, RNN-VSE, T-VSE) with an embedding size of $256$, using the following procedure and parameters.

\noindent
\textbf{CNN Model.} We used an ImageNet pre-trained DenseNet 169~\cite{densenet-cvpr17} with input image size $227 \times 227$ as the image encoder.
 	
\noindent
\textbf{Image Data Augmentation.} At training time, first resize to $333 \times 333$, then apply random crop of $227 \times 227$ and random horizontal flip with probability $0.5$. At test time, first resize to $333 \times 333$, then center crop.
 	
\noindent
\textbf{Two-stage Training.} Stage 1: Freeze the convolutional layers of the pre-trained CNN, train the embedding layer and all the text encoder from scratch, for $2$ epochs, with Adam optimizer, initial learning rate of $10^{-4}$, reduced by half after $1$ epoch. Stage 2: Train the whole VSE model for a maximum of $30$ epochs, with Adam optimizer, initial learning rate of $4 \times 10^{-5}$, reduced by half after $5$ and $10$ epochs. Evaluate the model on the validation set after each epoch and save the best model.

Note that freezing the pre-trained CNN in stage 1 is crucial, otherwise the model does not converge. Lastly, we trained the text encoders from scratch, since the new vocabulary invalidates all the weights of the pre-trained models.
	
\noindent
\textbf{Loss Function.} Symmetric triplet loss (max of hinges) with hard negative mining (max of hinges), with a margin value $m=0.2$, as described in Section~\ref{sec:loss}.
	
\noindent
\textbf{Batch Size.} We used a batch size of $256$ so that T-VSE model can fit on 4 NVIDIA V100 GPUs, each with 16GB memory, during training (at test time, a single V100 GPU is sufficient to compute both image and text embeddings concurrently with a batch size of 512).

\section{Experiments}
We trained all three models (T-VSE, AVG-VSE, RNN-VSE) on the training set of 11M product (image, title) pairs, evaluated on the 100K validation set, and tested the best model on the 1M test set.

For evaluation, we used the cosine distance to match each title/image to all the images/titles of the test set. As is common practice~\cite{vse-bmvc18}, we evaluated using R@K (Recall at K): the fraction of queries for which at least one correct result is returned in top K. We assume that for each query title/image, there is only one relevant image/title in the test set, although there are some duplicate and near-duplicate products in the dataset that hurt the performance.

\begin{table}[tb]	
	\centering
	\small
	\begin{tabular}{l l c c c c}
		\hline
		Model / R@K&  &  1 & 10 & 50 & 100 \\
		
		\hline
		\multirow{2}{*}{AVG-VSE}
			& t2i & 12.7  & 46.7 & 73.0 & 81.7 \\
			& i2t & 8.4 & 31.3 & 52.4 & 61.2 \\
		
		\hline 
		\multirow{2}{*}{RNN-VSE}
		    & t2i & 12.7 & 47.3 & 74.5 & 83.3 \\
		    & i2t & 8.1 & 29.6 & 50.9 & 60.9 \\
		
		\hline
		\multirow{2}{*}{T-VSE (6 layers)}
		   	& t2i & \textbf{30.7} & \textbf{76.6} & \textbf{91.9} & \textbf{95.1} \\
		    & i2t & \textbf{32.7} & \textbf{78.6} & \textbf{92.8} & \textbf{95.8} \\
		 	
		\hline
	\end{tabular}
	\caption{Text-to-image (t2i) and image-to-text (i2t) retrieval results on 1M test set of fashion product images-titles. Vocabulary size: 30K. Training epochs: 2+30.}
	\label{table:results}
\end{table}

\begin{table}[tb]	
	\centering
	\small
	\begin{tabular}{l l c c c c}
		\hline
		Model / R@K&  &  1 & 10 & 50 & 100 \\
		
		\hline
		\multirow{2}{*}{T-VSE (6 layers)}
		& t2i & 34.5 & 80.8 & 93.3 & 95.9 \\
		& i2t & 36.6 & 82.2 & 94.1 & 96.4 \\
		
		\hline
		\multirow{2}{*}{T-VSE (12 layers)}
		& t2i & \textbf{38.1} & \textbf{83.3} & \textbf{94.1} & \textbf{96.3} \\
		& i2t & \textbf{40.5} & \textbf{85.0} & \textbf{94.8} & \textbf{96.8} \\
		
		\hline
	\end{tabular}
	\caption{T-VSE with 6 and 12 transformer layers, larger batch size (400) and longer training (2+50 epochs).}
	\label{table:results2}
\end{table}

Table~\ref{table:results} presents the R@K accuracy on the 1M test set for the three VSE models, all trained for $2+30$ epochs (Sec.~\ref{sec:train}). T-VSE outperforms the other two models by a large margin on both text-to-image and image-to-text retrieval.

Table~\ref{table:results2} presents two more experiments that further improve the performance of the T-VSE model. (i) We further trained T-VSE for $20$ more epochs with a larger batch size of $400$.
We trained T-VSE with a 12-layer DistilBERT model for $2+50$ epochs with batch size 400. The results show that text model matters a lot in VSE, as well as the scale of the dataset. Transformer models can better leverage large training sets. They also benefit from longer training before the model saturates, as they have larger capacity.
	
AVG-VSE and RNN-VSE perform similarly, and image-to-text works worse than text-to-image, even though the loss function is symmetric. In T-VSE, image-to-text works slightly better than  text-to-image at all recall levels; probably due to the more powerful text model.

\section{Conclusions and Future Work}
We showed that properly trained transformer-based visual semantic embedding (T-VSE) models achieve vastly superior results on cross-modal image/text retrieval, compared to classical VSE models that employ RNNs or simple word average. We experimented with the DistilBERT model, but other transformer models should also work as well or even better.

Finding enough parallel image-text data to train these data ``hungry'' models is challenging, but the e-commerce sites have plenty of such data readily available. Hence, T-VSE training can also be used as a pre-training step for many other problems.
As a future work, it would also be interesting to explore unsupervised pre-training strategies for the transformer on product titles and descriptions, before the joint VSE training.

{\small
\bibliographystyle{ieee}
\bibliography{references}
}

\end{document}